\documentclass[10pt,twocolumn,letterpaper]{article}

\usepackage[pagenumbers]{cvpr} 

\usepackage{graphicx}
\usepackage{amsmath}
\usepackage{amssymb}
\usepackage{booktabs}
\usepackage{enumitem}

\usepackage{adjustbox}
\usepackage{ulem}
\usepackage{soul}
\usepackage[table]{xcolor}
\usepackage[symbol]{footmisc}
\usepackage{cuted}
\usepackage{capt-of}

\definecolor{tabfirst}{rgb}{1, 0.7, 0.7} 
\definecolor{tabsecond}{rgb}{1, 0.85, 0.7} 
\definecolor{tabthird}{rgb}{1, 1, 0.7} 

\usepackage[pagebackref,breaklinks,colorlinks]{hyperref}

\usepackage[capitalize]{cleveref}
\crefname{section}{Sec.}{Secs.}
\Crefname{section}{Section}{Sections}
\Crefname{table}{Table}{Tables}
\crefname{table}{Tab.}{Tabs.}

\begin{document}

\title{HexaGen3D: StableDiffusion is just one step away from\\ Fast and Diverse Text-to-3D Generation}

\author{Antoine Mercier $\hspace{0.5cm}$ Ramin Nakhli\thanks{Work done at Qualcomm AI Research during an internship.} $\hspace{0.5cm}$ Mahesh Reddy $\hspace{0.5cm}$ Rajeev Yasarla \\ Hong Cai $\hspace{0.5cm}$ Fatih Porikli $\hspace{0.5cm}$ Guillaume Berger\\
Qualcomm AI Research\thanks{Qualcomm AI Research is an initiative of Qualcomm Technologies, Inc.}\\
{\tt\small \{amercier, mahkri, ryasarla, hongcai, fporikli, guilberg\}@qti.qualcomm.com} \vspace{-4pt}
}
\maketitle
\begin{strip}
    \centering
    \includegraphics[width=0.9\textwidth]{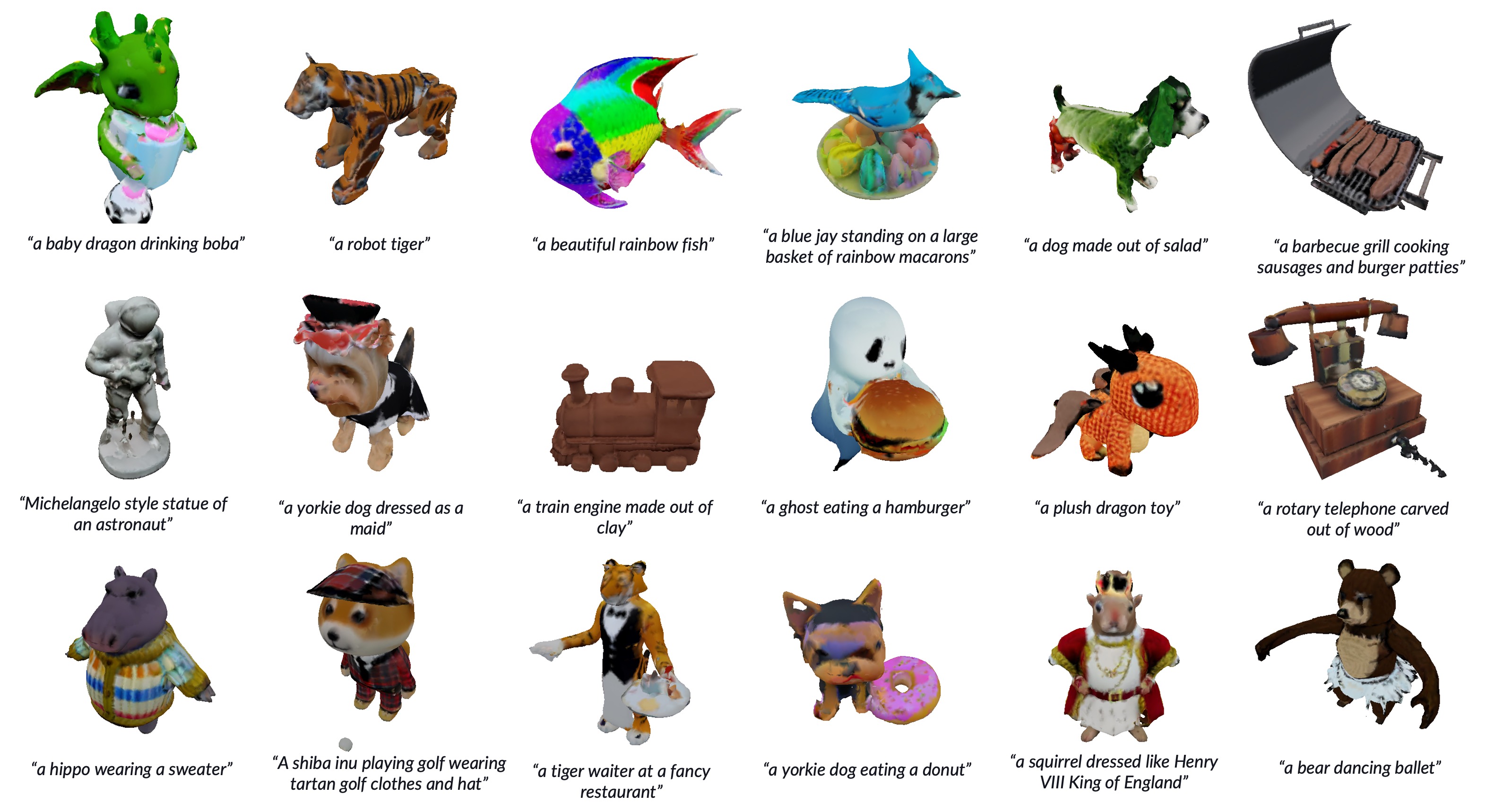}
    \captionof{figure}{HexaGen3D leverages pretrained text-to-image models to generate high-quality textured meshes in $\textbf{7}$ seconds. Our approach works well with a broad range of text prompts and generalize to new objects or object compositions not encountered during finetuning.}
    \label{fig:teaser}
\end{strip}

\begin{abstract}
\vspace{-3pt}
Despite the latest remarkable advances in generative modeling, efficient generation of high-quality 3D assets from textual prompts remains a difficult task. A key challenge lies in data scarcity: the most extensive 3D datasets encompass merely millions of assets, while their 2D counterparts contain billions of text-image pairs. To address this, we propose a novel approach which harnesses the power of large, pretrained 2D diffusion models. More specifically, our approach, HexaGen3D, fine-tunes a pretrained text-to-image model to jointly predict 6 orthographic projections and the corresponding latent triplane. We then decode these latents to generate a textured mesh. HexaGen3D does not require per-sample optimization, and can infer high-quality and diverse objects from textual prompts in 7 seconds, offering significantly better quality-to-latency trade-offs when comparing to existing approaches. Furthermore, HexaGen3D demonstrates strong generalization to new objects or compositions.

\end{abstract}

\section{Introduction}
\vspace{-3pt}

The creation of 3D assets, crucial across various domains such as gaming, AR/VR, graphic design, anime, and movies, is notably laborious and time-consuming. Despite the use of machine learning techniques to aid 3D artists in some aspects of the task, the development of efficient end-to-end 3D asset generation remains challenging, primarily bottlenecked by the limited availability of high-quality 3D data.

In contrast, image generation has recently witnessed a surge in quality and diversity, driven by the training of large models \cite{ramesh2021zero, yu2022scaling, rombach2022high, saharia2022photorealistic, podell2023sdxl, kang2023scaling} on large datasets like LAION-5B \cite{schuhmann2022laion} containing billions of text-image pairs. This progress in 2D generative modeling, coupled with the lack of curated 3D data, has prompted recent efforts to adapt 2D pre-trained models for 3D asset creation. One notable example is DreamFusion \cite{poole2022dreamfusion}, which utilizes the prior knowledge of a large 2D diffusion model to guide the optimization of a 3D representation, usually a NeRF \cite{mildenhall2021nerf}. While this approach, dubbed Score Distillation Sampling (SDS), and follow-up variants have showed promising results, their main limitation remains the extremely long generation time, taking anywhere from 20 minutes to several hours, and the lack of diversity across seeds.

To address these challenges, we present HexaGen3D, a novel text-to-3D model that significantly reduces generation time without sacrificing quality or diversity. Similar to DreamFusion, we build upon pre-trained 2D models but instead propose to modify and finetune them for direct, feed-forward, generation of 3D objects. Below are the key contributions of our work:

\begin{itemize}

    \item We adopt a similar approach to 3DGen \cite{gupta20233dgen}, but replace their custom-designed latent generative model with a pre-trained text-to-image model, reducing the need for extensive 3D finetuning data and enabling generalization to new objects or compositions not encountered during training.

    \item We introduce ``Orthographic Hexaview guidance", a novel technique to align the model's 2D prior knowledge with 3D spatial reasoning. This intermediary task involves predicting six-sided orthographic projections which we then map to our final 3D representation. This allows the U-Net of existing 2D diffusion models to efficiently perform multi-view prediction and 3D asset generation in a sequential manner, with 3D generation only requiring one additional U-Net inference step.

    \item HexaGen3D competes favorably with existing approaches in quality while taking only seven seconds on an A100 to generate a new object. This speed is orders of magnitude faster than existing approaches based on SDS optimization.

\end{itemize}

\begin{figure*}[ht!]
    \centering
    \begin{subfigure}[t]{0.545\textwidth}
        \centering
        \caption{HexaGen3D Overview}
        \label{fig:method_overview_hexagen3d_panel}\includegraphics[width=\linewidth]{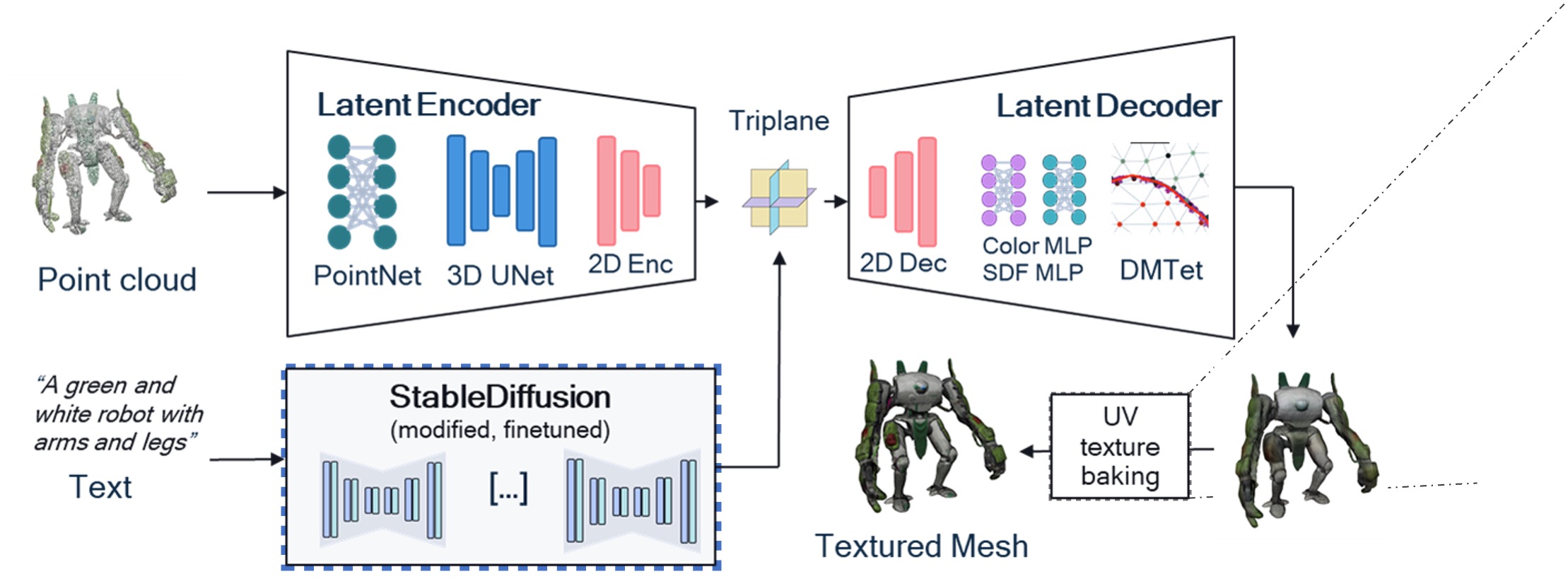}
    \end{subfigure}
    \vspace{-0.3pc}
    \hspace{-2pc}
    \begin{subfigure}[t]{0.41\textwidth}
        \centering
        \caption{UV Texture Baking} \label{fig:method_overview_texture_baking_panel}\includegraphics[width=\linewidth]{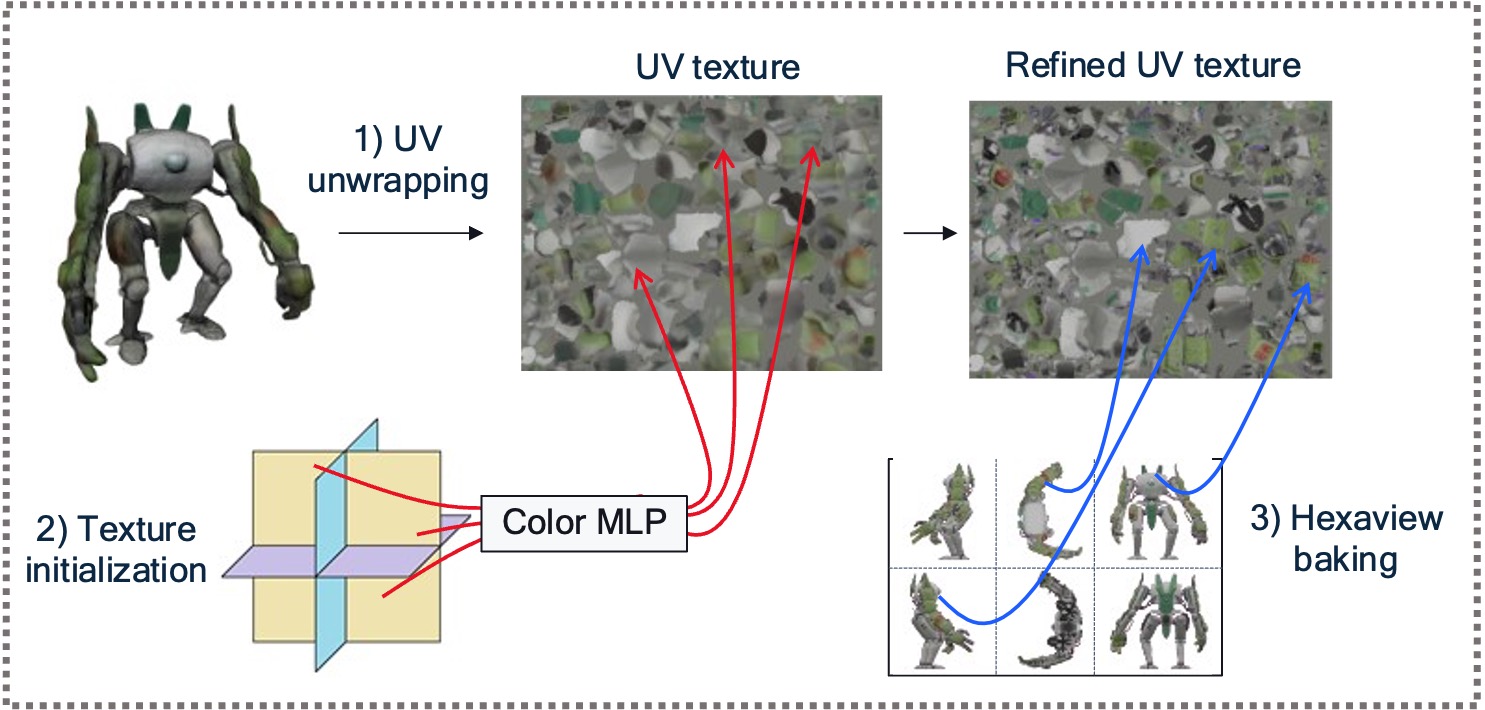}
    \end{subfigure}
    \begin{subfigure}[t]{0.545\textwidth}
        \centering
        \includegraphics[width=\linewidth]{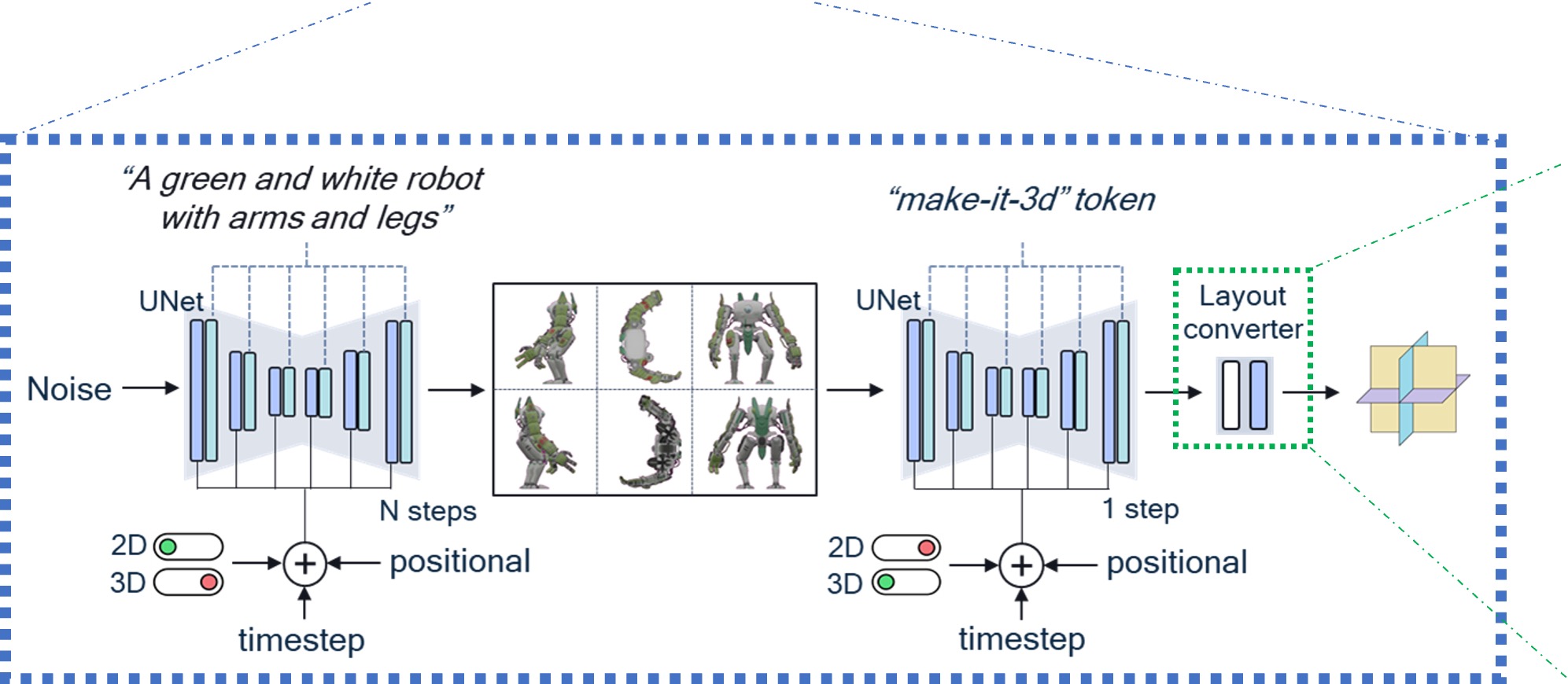}
        \caption{Triplanar Latent Generation} \label{fig:method_overview_triplanar_generation_panel}
    \end{subfigure}
    \hspace{-0.5pc}
    \begin{subfigure}[t]{0.445\textwidth}
        \centering
        \includegraphics[width=\linewidth]{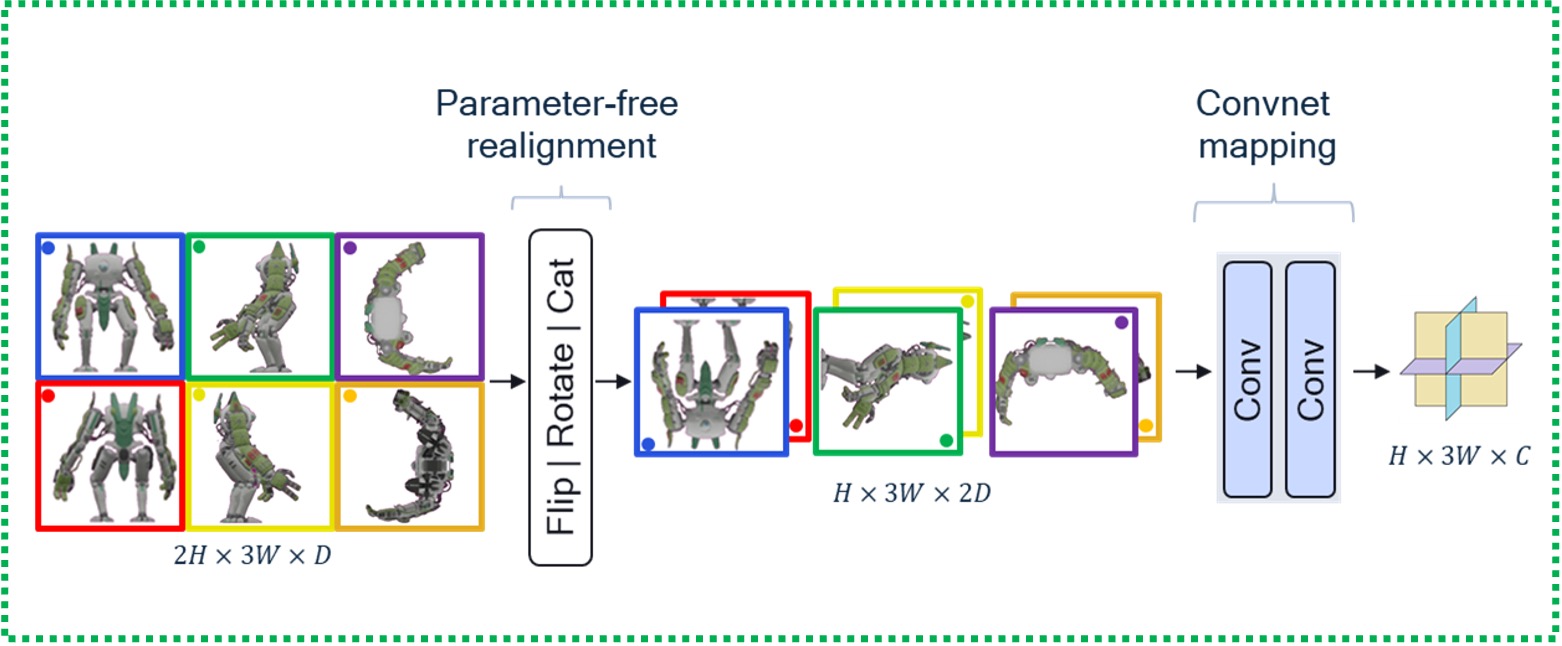}
        \caption{Hexa-to-triplane layout conversion} \label{fig:method_overview_hexa_to_triplane_panel}
    \end{subfigure}
    \caption{Method overview. HexaGen3D predicts textured meshes from textual prompts in 7 seconds, leveraging pre-trained text-to-image models for latent generation. Like 3DGen \cite{gupta20233dgen}, the training procedure comprises two stages and the latent representation is based on triplanes. Top-left (a): High-level overview of the two stages: triplanar latent auto-encoder and StableDiffusion-based latent generation.  Bottom-left (c): StableDiffusion-based triplanar latent generation via orthographic hexaview diffusion (section \ref{sec:latent-diffusion}). Bottom-right (d): Detailed overview of our hexa-to-triplane layout converter (section \ref{sec:hexa-to-triplane-realignment}). Top-right (b): Detailed overview of our UV texture baking procedure, used as a post-processing step to improve the visual appearance of the decoded mesh (section \ref{sec:texture-baking}).}
    \label{fig:method_overview}
\end{figure*}

\section{Related Work}
Recent methods heavily leverage diffusion models~\cite{ho2020denoising} for generating 3D meshes from single images and/or textual prompts. In this section, we briefly review these works and refer readers to~\cite{po2023state} for a more comprehensive survey.

\textbf{Using Score Distillation Sampling:} Pioneering works, such as DreamFusion~\cite{poole2022dreamfusion} and Latent-NeRF~\cite{metzer2023latent} generates 3D by training a NeRF using Score Distillation Sampling (SDS). However, they require lengthy, per-sample optimization, taking $\sim$2 hours to generate an object.
Later methods accelerate this process by first producing a coarse NeRF model and then upscaling or enhancing it, e.g., Magic3D~\cite{lin2023magic3d}, Make-it-3D~\cite{tang2023make}, or by first generating geometry and then baking the texture using images generated from diffusion~\cite{tsalicoglou2023textmesh}. These techniques considerably reduce generation time, e.g., to 20 - 40 minutes. More recently, MVDream~\cite{shi2023mvdream} replaces the guidance model with a multi-view diffusion model, leading to improved visual quality, at the cost of slower generation times, taking up to 3 hours to generate an object. ATT3D~\cite{lorraine2023att3d} proposes to map text prompts to a NeRF embedding using a network. This enables generating a mesh within a second, but lacks mesh quality and generalizability.

\textbf{Feedforward Generation via Diffusion:}
Several works use diffusion to directly generate 3D structure or representation~\cite{lorraine2023att3d,nichol2022point,gupta20233dgen,luo2021diffusion, zhang20233dshape2vecset, shue20233d, zhao2023michelangelo, kim2023neuralfield,chen2023single}. Point-E~\cite{nichol2022point} uses diffusion to generate a point cloud, conditioning on an image synthesized from a text input. Shap-E~\cite{jun2023shap} uses a similar transformer-based diffusion architecture but generates parameters of implicit 3D functions like NeRF or DMTet. More recently, 3DGen~\cite{gupta20233dgen} adopts a latent triplane representation, which can be denoised by a diffusion model and fed to a point cloud VAE decoder to generate a 3D mesh. SSDNeRF~\cite{chen2023single} also adopts the triplane representation but uses a NeRF decoder.
\textbf{Leveraging 2D View Synthesis:}
These works utilize diffusion models to generate new 2D views to support 3D generation.  3DiM~\cite{watson2022novel} generates multiple views of target poses, based on a source view and its pose. One-2-3-45~\cite{liu2023one} utilizes Zero-1-to-3~\cite{liu2023zero1to3} to generate several views of an object. It then constructs a volumetric representation based on the generated images and can produce a mesh within 45 seconds. However, the 2D views are generated one at a time and may not provide consistent multi-view information for the volume.
Other works employ diffusion to generate multiple views simultaneously, e.g., SyncDreamer~\cite{liu2023syncdreamer}, and Instant3D~\cite{instant3d2023}, and reconstructs 3D objects based on these synthesized views.
Wonder3D~\cite{long2023wonder3d} creates a multi-domain diffusion model to generate both 2D images and normal maps, and trains a signed distance function to extract the underlying textured mesh.
Some papers further introduce 3D awareness into the image synthesis process, e.g., GeNVS~\cite{chan2023generative}, NerfDiff~\cite{gu2023nerfdiff}, SparseFusion~\cite{zhou2023sparsefusion}, Sparse3D~\cite{zou2023sparse3d}, providing better multi-view consistency.

\textbf{Texture Baking:} While some of the aforementioned works can also be used to refine textures of an existing mesh~\cite{poole2022dreamfusion,lin2023magic3d}, several studies focus on texture baking.
For instance, Text2Tex~\cite{chen2023text2tex} and TEXTure~\cite{richardson2023texture} uses off-the-shelf StableDiffusion models \cite{rombach2022high} to iteratively paint texture over known geometry. X-Mesh~\cite{ma2023x} adopts a dynamic attention module and the CLIP loss between the text prompt and rendered images to generate texture. TexFusion~\cite{cao2023texfusion} applies a diffusion model's denoiser on rendered 2D images and aggregates the denoising predictions on a shared latent texture map to achieve better texture consistency.

\section{Method}

Our latent generative approach, HexaGen3D, leverages pre-trained text-to-image models to predict textured meshes from textual prompts. Like 3DGen \cite{gupta20233dgen}, the training procedure comprises two stages. In the first stage, we learn a triplanar representation --- capturing the shape and color of a textured mesh --- using a variational auto-encoder (VAE) \cite{kingma2019introduction}. In the second stage, we finetune a pre-trained text-to-image model to synthesize new samples in this triplanar latent space. Figure \ref{fig:method_overview_hexagen3d_panel} provides a comprehensive overview of each component.

At test time, we infer new 3D assets in a feedforward manner --- without requiring per-sample optimization --- allowing for faster and more efficient generation. We first produce triplanar latents from random noise using our latent generator and then convert these latents into a textured mesh using the decoder from the first stage.

For stage 1, our approach retains most architecture components of the auto-encoder from 3DGen \cite{gupta20233dgen} (detailed in section \ref{sec:triplane-autoencoder}). However, a key distinction of our work is that we propose to leverage pre-trained text-to-image models for stage 2 (detailed in section \ref{sec:latent-diffusion}). Notably, our results demonstrate that, with proper guidance and minimal finetuning data, 2D generative models such as StableDiffusion transfer surprisingly well to 3D asset generation.

\subsection{Stage 1: Learning the Triplanar Tepresentation}
\label{sec:triplane-autoencoder}


In this stage, we train a VAE to learn a triplanar latent representation of textured meshes, derived from point clouds. Our solution closely follows the framework established by 3DGen \cite{gupta20233dgen}, as we re-use several components from their methodology. Initially, a PointNet model \cite{Qi_2017_CVPR} encodes the input point cloud into a triplanar representation. These latents are subsequently refined through two successive networks, and then converted into a textured mesh using a decoder based on the DMTet algorithm \cite{shen2021deep}, as depicted in \Cref{fig:method_overview_hexagen3d_panel}.

Similar to 3DGen, our pipeline utilizes a U-Net with 3D-aware convs \cite{wang2023rodin} for the first refinement network. However, for the second network, we replace the 3D-aware U-Net architecture used in 3DGen to the VAE architecture typically used in StableDiffusion models \cite{rombach2022high} and utilize the low-resolution intermediate features from this second network as the target representation for latent generation.

\subsection{Stage 2: Triplanar Latent Generation using Pretrained 2D Diffusion Models}
\label{sec:latent-diffusion}

In this stage, our goal is to generate triplanar latents that are consistent with the provided textual prompt. Unlike approaches such as 3DGen~\cite{gupta20233dgen} or Shap-E~\cite{jun2023shap} which focus on developing and training novel architectures from scratch, we aim to build upon pre-trained text-to-image models and explore how to extend them to the 3D domain. A key aspect of our approach is the introduction of a novel intermediary task, ``Orthographic Hexaview" prediction, specifically designed to bridge the gap between 2D and 3D synthesis.

\subsubsection{Orthographic Hexaview Guidance}

As typically done in the literature, our triplanar representation consists of three feature maps concatenated along the width axis to form a $1 \times 3$ ``rolled-out" layout. This two-dimensional arrangement makes it possible to apply 2D architectures to triplanar latent generation, yet this task still requires 3D geometric reasoning capabilities to ensure feature consistency across the three planes. Our results suggest that pre-trained text-to-image models, despite their prior knowledge on how to arrange multiple objects coherently, struggle when directly fine-tuned to generate such rolled-out triplanes, potentially due to the limited 3D data available during fine-tuning. To address this, we decompose the generation process into two steps, introducing an intermediate ``hexaview'' representation designed to guide the latent generation process.

\begin{figure}[t!]
\begin{center}
   \includegraphics[width=\linewidth]{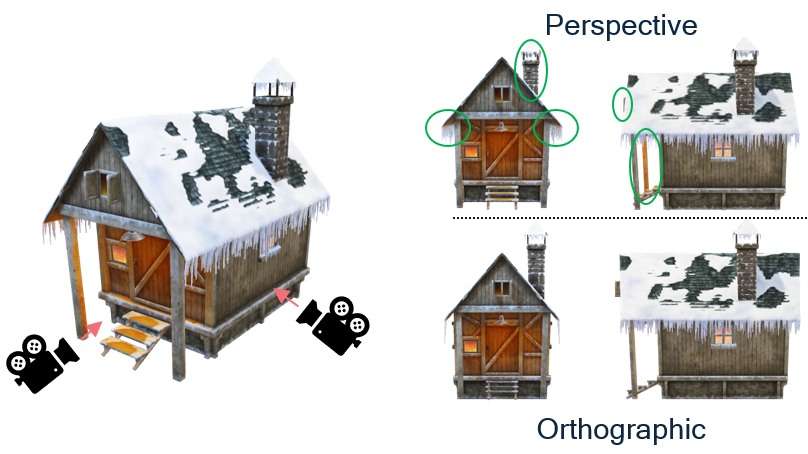}
\caption{Perspective vs Orthographic views. Orthographic views are free from perspective distortions (highlighted in green), facilitating re-alignment to triplanar representations.}
\label{ortho-vs-perspective}
\end{center}
\end{figure}

\begin{figure}[t!]
    \centering
        \includegraphics[width=\columnwidth]{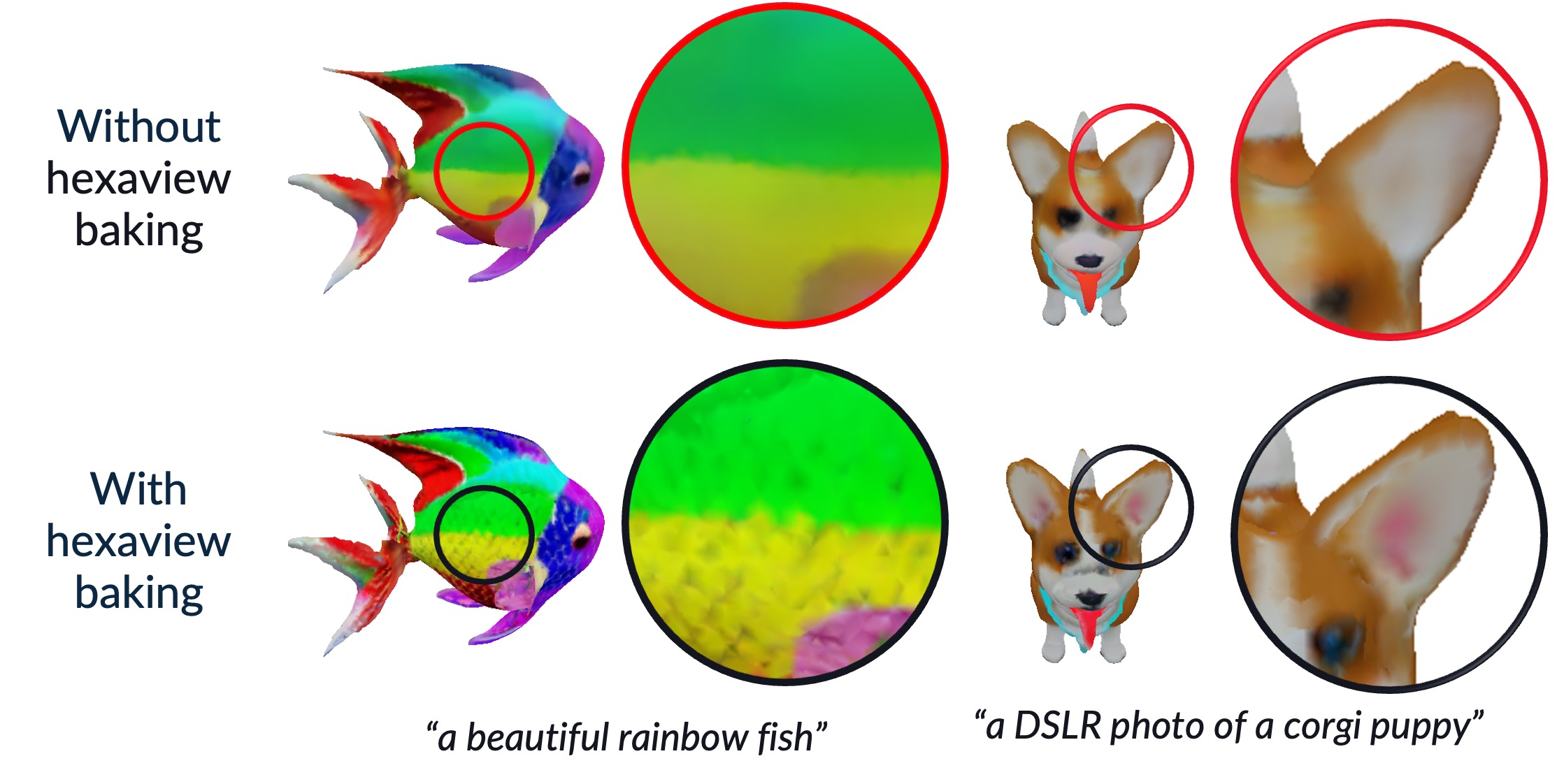}
        \caption{
        The orthographic hexaview prediction contains realistic details in the pixel space compared to the generated mesh. To significantly improve the visual feel of the final generated mesh, we ``bake'' the detailed hexaview prediction to the UV texture map (see Sec.~\ref{sec:texture-baking}).
        }
    \label{fig:qualt_comparison_textures}
\end{figure}

As depicted in \cref{fig:method_overview_triplanar_generation_panel}, the intermediate ``hexaview" representation includes latent features from six orthogonal views (front, rear, right, left, top, and bottom), spatially concatenated to form a $2 \times 3$ layout. The triplanar latent generation process is decomposed into two stages. We first diffuse hexaview latents using a modified and finetuned text-to-image U-Net (detailed in section \ref{sec:unet-modif}). Then, we map these latents to our target triplanar representation, re-using the same diffusion U-Net for an additional inference step and a small layout converter (detailed in section \ref{sec:hexa-to-triplane-realignment}). Computation-wise, this means we perform $N + 1$ U-Net inferences in total, $N$ steps to obtain a multi-view representation of the object, and $1$ additional U-Net forward pass to generate the corresponding mesh.

Unlike concurrent 3D asset generation approaches based on multi-view prediction \cite{shi2023mvdream, instant3d2023}, a key and novel feature of our method is the use of orthographic projections\footnote{\tiny\url{https://en.wikipedia.org/wiki/Orthographic_projection}}. As shown in \cref{ortho-vs-perspective}, this form of parallel projection, commonly used in mechanical drawings of 3D parts, avoids perspective distortions, thereby guaranteeing pixel-accurate re-alignment of each orthogonal view to the corresponding plane in our target latent representation.

Another novel aspect of our approach is that the same 2D U-Net is trained to perform multi-view prediction and 3D asset generation simultaneously. To the best of our knowledge, this multi-task scenario has never been explored before. Interestingly, our results show that the same network can perform both tasks well and that the proposed hexaview guidance is in fact instrumental in unlocking high-quality 3D asset generation, offering a new technique to capitalize on the synthesis capabilities of large pre-trained text-to-image models.

\subsubsection{U-Net architecture modifications}
\label{sec:unet-modif}


In our experiments, we adapt the U-Net architecture from StableDiffusion models, though we believe our method could be effectively applied to other text-to-image models as well. \cref{fig:method_overview_triplanar_generation_panel} illustrates the modifications we applied to the U-Net architecture, aimed at enhancing its suitability for the 3D domain:

\paragraph{Additional conditioning signals:}  Our first modification introduces two types of learnable encodings which we add to the timestep embeddings, providing a mechanism to modify the timestep-based modulations of convolutional and attention layers performed in the original U-Net. This modification draws inspiration from techniques like the camera embedding residuals in MVDream \cite{shi2023mvdream} or the domain switcher proposed in Wonder3D \cite{long2023wonder3d}. The first type of encodings is based on positional embeddings. These help in meeting the strict spatial alignment constraints imposed by the hexaview and triplane representations and can be thought as serving the same purpose as the 3D-aware convolutions in 3DGen~\cite{gupta20233dgen}. The second type, domain encodings, involves learning distinct feature vectors for each phase of the generation process – one for the $N$ hexaview diffusion steps (2D domain) and another for the final triplane mapping step (3D domain). This additional encoding allows our model to adjust its processing dynamically based on the current stage of the latent generation process.

\paragraph{``Make-it-3d" token:} Another modification is the introduction of a special ``Make-it-3d" token during the hexaview-to-triplanar mapping step. The corresponding embedding vector is attended to via cross-attention layers and gives the U-Net component yet another mechanism to adapt its behavior specifically for triplanar latent prediction.

\subsubsection{Hexa-to-triplane Layout Converter}
\label{sec:hexa-to-triplane-realignment}

Following hexaview diffusion, we apply the U-Net for an additional inference step, using the 3D-specific conditioning techniques discussed in the previous section, and extract its penultimate features. We then map these features to the target triplanar representation using a specialized module called hexa-to-triplane layout converter. As shown in \cref{fig:method_overview_hexa_to_triplane_panel}, our layout converter starts by re-orienting each view from the $2 \times 3$ input layout to align it with the corresponding plane from our 3D latent representation. This is achieved through a series of parameter-free operations, including slicing to isolate individual views, rotating and flipping to orient them correctly, and concatenating parallel views (\ie top and bottom, left and right, front and back). The use of orthographic projections is advantageous here, as it guarantees pixel-accurate layout conversion and eliminates perspective distortions. A convolutional neural network (CNN) then predicts the final triplanar latents. In practice, the CNN's main purpose is to adjust the number of channels to align with our independently-trained VAE. Our experiments indicate that the U-Net is effective in multi-tasking and predicting both hexaview latents and 3D-conditioned features that map well to the final triplanar latents. Consequently, the additional CNN does not need to be very complex.

\begin{figure*}[!t]
    \centering
        \includegraphics[width=0.9\textwidth]{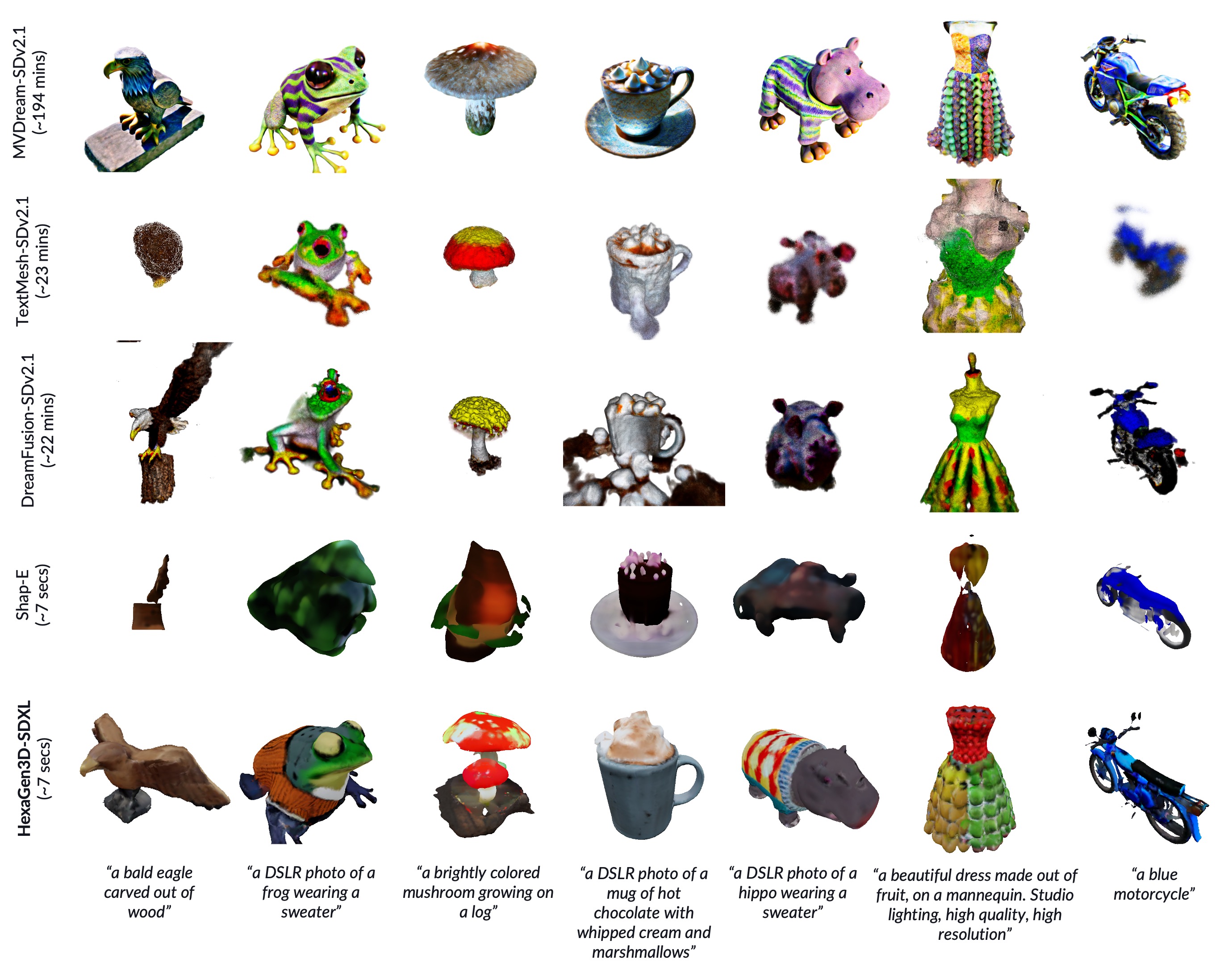}
        \caption{We compare ``HexaGen3D'' with other text-to-3D approaches: DreamFusion-SDv2.1 (DF)~\cite{poole2022dreamfusion}, TextMesh-SDv2.1 (TM)~\cite{tsalicoglou2023textmesh}, MVDream-SDv2 (MV)~\cite{shi2023mvdream}, and Shap-E~\cite{jun2023shap}. We use the threestudio~\cite{threestudio2023} framework that implements the DF, TM, and MV approaches that leverages the StableDiffusion~\cite{rombach2022high} text-to-image model. We use the text prompts from threestudio~\cite{threestudio2023} based on DF.}

    \label{fig:qualt_comparison_text_to_3d}
\end{figure*}
\subsection{Texture baking for better visual appearance}
\label{sec:texture-baking}

In order to further enhance the visual appearance of the final mesh, we introduce a UV texture baking approach that leverages the six-sided orthographic views predicted during hexaview diffusion. Interestingly, these views often contain fine textural details that are missing in the VAE-decoded mesh. To address this, we adopt a post-processing procedure which ``bakes" the intermediate hexaviews onto the decoded mesh, as illustrated in \cref{fig:method_overview_texture_baking_panel}. This procedure involves three steps:

\begin{enumerate}
    \item \textbf{UV Mapping:} We employ an automatic UV unwrapping algorithm to create a UV mapping, making it possible to project a 2D texture onto the 3D mesh.
    \item \textbf{Texture Initialization:} We then initialize our UV texture, mapping each texel to its corresponding 3D coordinate, sampling the corresponding triplanar feature, and converting it to RGB using the Color MLP from our VAE decoder. While this step alone does not substantially enhance visual quality --- even with increased texture resolution --- it is crucial for areas not fully covered by the final hexaview baking step.
    \item \textbf{Hexaview Baking:} We project pixels from the six (decoded) orthographic views onto the UV texture. In cases where multiple hexaview pixels correspond to the same texel, we average their values. This final step significantly enhances the visual quality of the mesh, as shown in \cref{fig:qualt_comparison_textures}.
\end{enumerate}

While this texture baking procedure effectively addresses the limitations posed by the initial mesh decoding, further improvements in the VAE pipeline represent a promising avenue for future research. We believe that refining this component, such as potentially using higher-resolution latents, could alleviate some of the current bottlenecks, and we reserve this exploration for future work.

\subsection{Implementation details}

\paragraph{Training data.} We train our models on 3D assets from the Objaverse dataset \cite{deitke2023objaverse}, complemented with automated captions generated by Cap3D \cite{luo2023scalable}. Following findings from prior work \cite{gupta20233dgen,shi2023mvdream,instant3d2023}, we filter out lower-quality assets from the original dataset, ultimately training our models on $86,784$ curated objects. The selection criteria for these assets were based on factors such as geometric complexity and textural richness (more details can be found in the appendix).

In addition to 3D assets, we augment our training set with regular images while finetuning the latent diffusion model. 
The incorporation of 2D data helps to mitigate the limited amount of available 3D data and enhances the model's generalization to unseen 3D objects. This form of multi-task regularization has been similarly employed in MVDream \cite{shi2023mvdream}.

In terms of losses, we follow 3DGen and combine a mask silhouette loss, a depth loss, a laplacian smoothness loss and a KL divergence loss to supervise the geometry VAE, $L_{geometry} = \alpha L_{mask} + \phi L_{depth} + \lambda L_{smooth} - \gamma D_{KL}$,  with $\alpha = 3$ , $\phi=10$, $\lambda=0.01$, and $\gamma=10^{-7}$, and use a sum of L1 and L2 losses to supervise the color VAE.

We train our VAE models for $15$ epochs, using a batch size of $16$, an initial learning rate of $3\times10^{-5}$ and a cosine annealing schedule down to a minimum learning rate of $10^{-6}$. This takes roughly a week on 8 A100s.

\begin{table*}[!t]
    \centering
    \begin{adjustbox}{}
        \small
        {\renewcommand{\arraystretch}{1.1}
            \begin{tabular}{l|c|ccc|cc}
            \hline
            \textbf{Method} & \multicolumn{1}{c|}{\textbf{Latency $\downarrow$}} & \multicolumn{3}{c|}{\textbf{CLIP score} $\uparrow$} & \multicolumn{2}{c}{\textbf{User preference score $\uparrow$}}\\
            \multicolumn{1}{c|}{} & Single A100 GPU & CLIP L/14 & CLIP B/16 & CLIP B/32 &  Vis. qual. & Prompt fid.
            \\ \hline
            
            MVDream-SDv2.1~\cite{shi2023mvdream}
                        & $\approx194$ mins
                        & \cellcolor{tabfirst} 25.02 
                        & \cellcolor{tabfirst} 30.35 
                        & \cellcolor{tabfirst} 29.61 
                        & \cellcolor{tabfirst} 0.97
                        & \cellcolor{tabfirst} 0.88
                        \\
            TextMesh-SDv2.1~\cite{tsalicoglou2023textmesh}
                         & \cellcolor{tabthird}$\approx23$ mins
                         & 19.46 
                         & 25.06 
                         & 24.86 
                         & 0.12
                         & 0.21
                         \\
            DreamFusion-SDv2.1~\cite{poole2022dreamfusion}
                            & \cellcolor{tabsecond}$\approx22$ mins
                            & 23.76 
                            & \cellcolor{tabthird} 28.91 
                            & \cellcolor{tabthird} 28.96 
                            & 0.50
                            & \cellcolor{tabthird}0.52
                            \\
            Shape-E~\cite{jun2023shap}
                        & \cellcolor{tabfirst} $\approx7$ secs
                        & 19.52 
                        & 24.33 
                        & 24.70 
                        & 0.17
                        & 0.13
                        \\
            \hline
            HexaGen3D-SDv1.5 
                        & \cellcolor{tabfirst} $\approx7$ secs 
                        & \cellcolor{tabthird}24.02 
                        & 28.84 
                        & 28.55 
                        & \cellcolor{tabthird}0.51
                        & 0.49
                        \\ 
            HexaGen3D-SDXL 
                        & \cellcolor{tabfirst}$\approx7$ secs 
                        & \cellcolor{tabsecond} 24.98 
                        & \cellcolor{tabsecond} 29.58 
                        & \cellcolor{tabsecond}  28.97 
                        & \cellcolor{tabsecond}0.73
                        & \cellcolor{tabsecond}0.77
                        \\
            \hline
            \end{tabular}
        }
    \end{adjustbox}
    \caption{Quantitative comparison of HexaGen3D with other approaches on 67 prompts from the DreamFusion~\cite{poole2022dreamfusion} prompts available in threestudio~\cite{threestudio2023}. We report the inference time on a single A100, CLIP scores for three different CLIP models, as well as user preference scores based on two criteria, visual quality and prompt fidelity. The preference scores reported here are obtained by averaging the pair-wise preference rates from \cref{fig:user_study}. 
    }
    \label{tab:quant_comparison_text_to_3d}
\end{table*}

\begin{figure}[!t]
    \centering
    \begin{subfigure}[t]{0.5\columnwidth}
        \centering
        \includegraphics[height=1.3in]{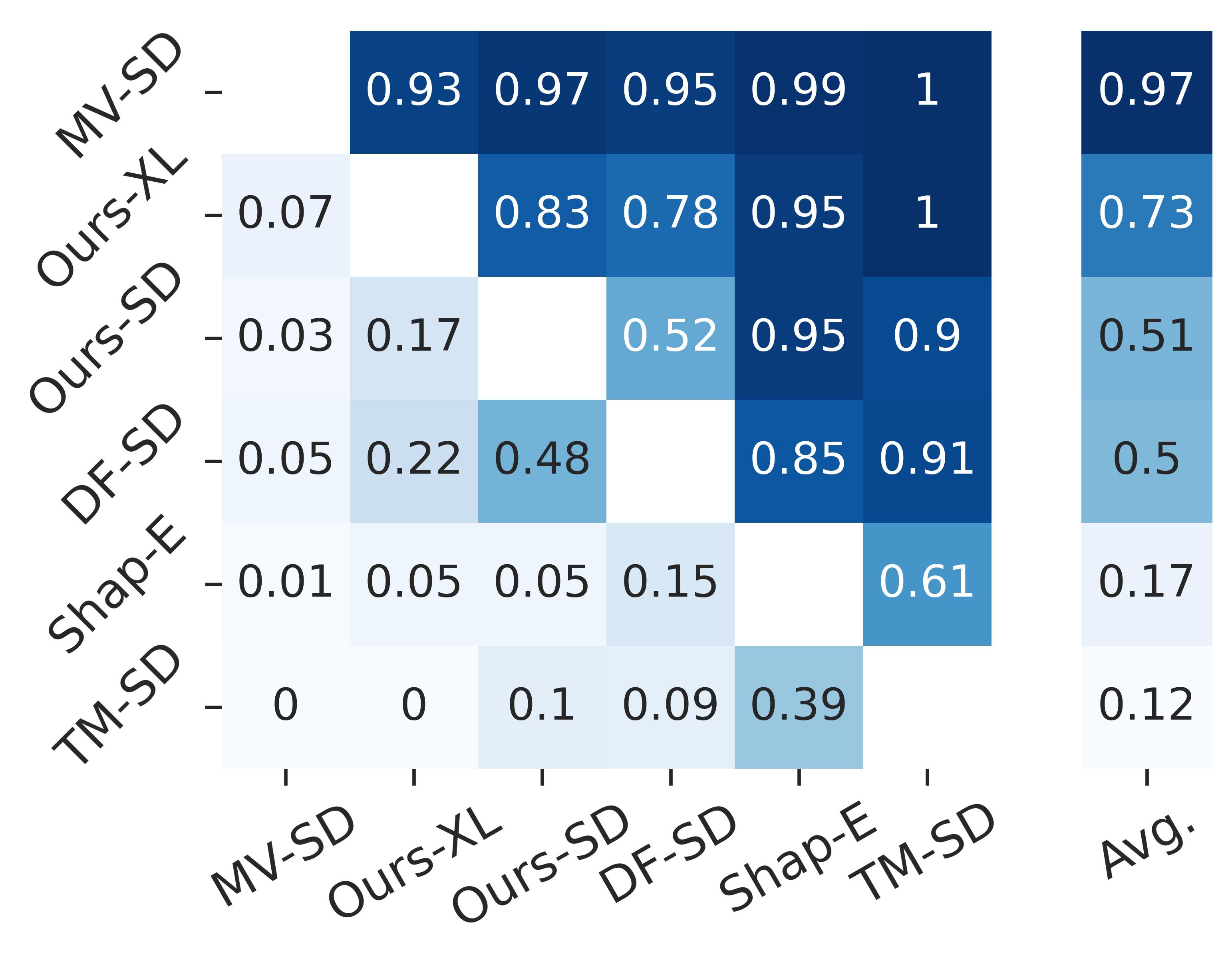}
        \caption{Evaluating the approaches for ``visual quality''.}
        \label{fig:user_study_1}
    \end{subfigure}%
    ~
    \begin{subfigure}[t]{0.5\columnwidth}
        \centering
        \includegraphics[height=1.3in]{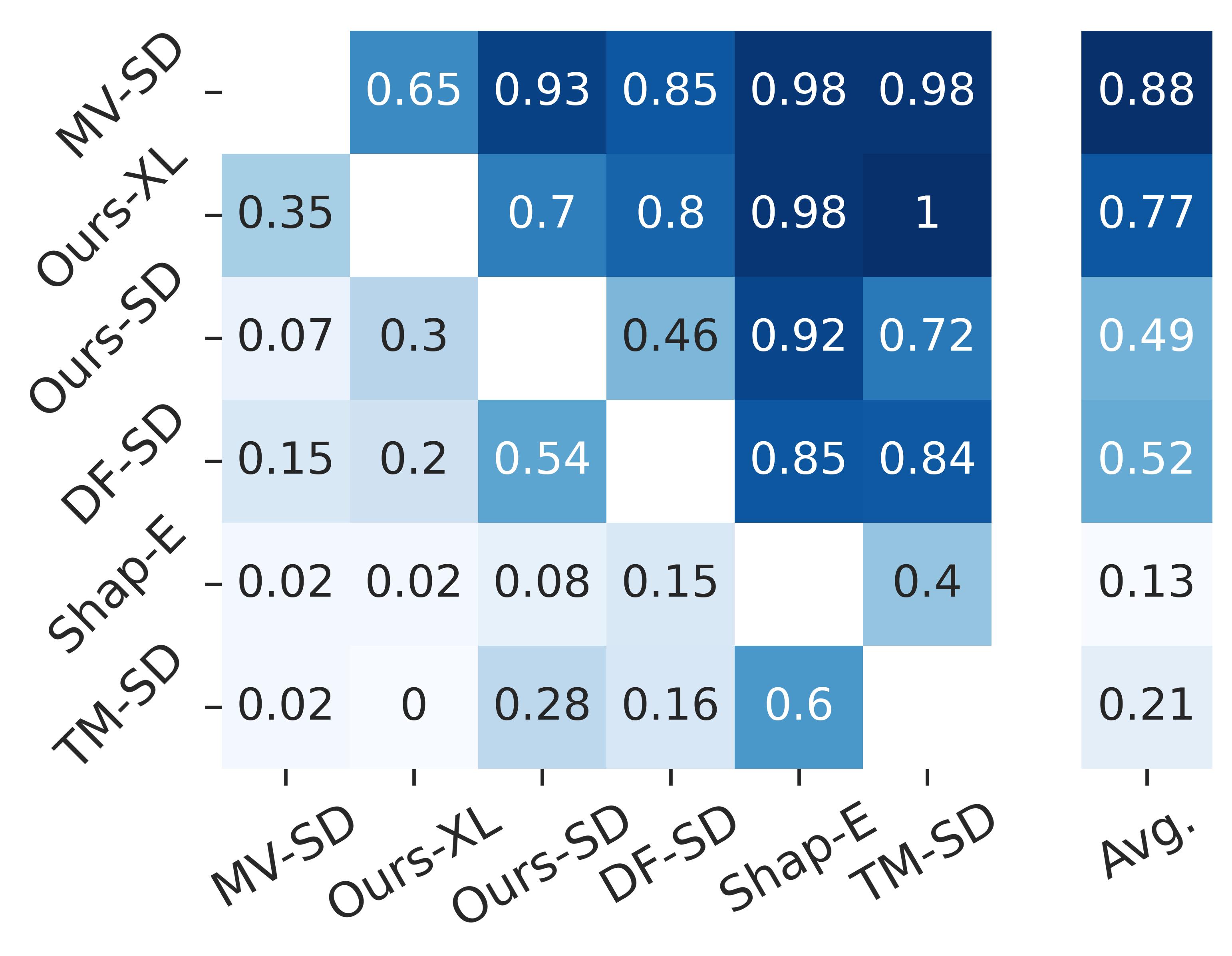}
        \caption{Evaluating the approaches on ``text prompt fidelity''.}
        \label{fig:user_study_2}
    \end{subfigure}
    \caption{User study comparing all the text-to-3D approaches on (a) visual quality and (b) text prompt fidelity. Each cell indicates the user preference score (\%) for an approach (row) over another (column). The approaches are: MVDream-SDv2.1 (MV-SD), DreamFusion-SDv2.1 (DF-SD), Shape-E, TextMesh-SDv2.1 (TM-SD), HexaGen3D-SDv1.5 (Ours-SD), and HexaGen3D-SDXL (Ours-XL).}
    \label{fig:user_study}
\end{figure}

\paragraph{Stage 2: Triplanar Latent Generation.} Architecture-wise, we experimented with two pretrained StableDiffusion U-Net backbones: SD-v1.5 \cite{rombach2022high} and SD-XL \cite{podell2023sdxl}. The neural net used within our hexa-to-triplane layout converter is a two-layer convnet with SiLU activations and trained from scratch.

For the intermediate hexaview generation, we rasterize six $512 \times 512$ orthographic projections of the ground truth mesh, downscale them to $256 \times 256$, and encode these images using the VAE encoder from the corresponding Stable Diffusion model. The resulting $32 \times 32$ latents are then re-organized into a $2 \times 3$ layout of size $64 \times 96$. For 2D regularization, we use high-res images which we downsize to $256 \times 256$, matching the resolution of individual views within our hexaview representation. Considering that StableDiffusion models often have difficulties with predicting plain white backgrounds \cite{instant3d2023}, we add random colored backgrounds to the hexaviews. This addition does not impact triplanar latent generation, as the model effectively learns to disregard the background color in this context.

During training, we effectively optimize the model to solve three tasks simultaneously: hexaview latent generation, hexaview-to-triplane mapping and image latent generation (for regularization). We achieve this using gradient accumulation, aggregating gradients from all three tasks. We use a batch of 192 hexaviews for the hexaview diffusion and hexaview-to-triplanar mapping tasks, and a batch of 32 regular images for 2D regularization. We down-weigh the diffusion loss coming from the 2D regularization batches by a factor $0.25$. We train our latent generative models for $50,000$ iterations using a learning rate of $3\times 10^{-5}$. This takes roughly 4 days on 8 A100s.

\paragraph{Baselines} We select a range of recent text-to-3D approaches for comparison with HexaGen3D, including a feedforward approach, Shap-E~\cite{jun2023shap}, and three SDS-based approaches,  DreamFusion~\cite{poole2022dreamfusion}, TextMesh~\cite{tsalicoglou2023textmesh}, and MVDream~\cite{shi2023mvdream}. While we use the official implementation of Shap-E\footnote{https://github.com/openai/shap-e} and MVDream\footnote{https://github.com/bytedance/MVDream-threestudio}, we leverage the DreamFusion and TextMesh implementations available within the threestudio framework~\cite{threestudio2023}, which include some notable deviations from their original papers. Specifically, our DreamFusion setup uses the open-source StableDiffusion v2.1 as the guidance model instead of Imagen~\cite{saharia2022photorealistic}, and a hash grid ~\cite{muller2022instant} as the 3D representation, diverging from MipNerf360~\cite{barron2022mip}. TextMesh similarly utilizes StableDiffusion v2.1 for guidance, with NeuS~\cite{wang2021neus} replacing VolSDF~\cite{yariv2021volume} for the 3D signed distance field representation.

\section{Experiments}

As illustrated in \cref{fig:teaser}, HexaGen3D generates high-quality textured meshes from a wide array of textual prompts, including complex prompts that involve unlikely combinations of objects, diverse styles or unique material properties. In \Cref{sec:comparison-against-existing-work}, we compare HexaGen3D with existing methods, and \Cref{sec:ablation_study} details our ablation study results.

\begin{figure}[t!]
    \centering
        \includegraphics[width=1.05\columnwidth]{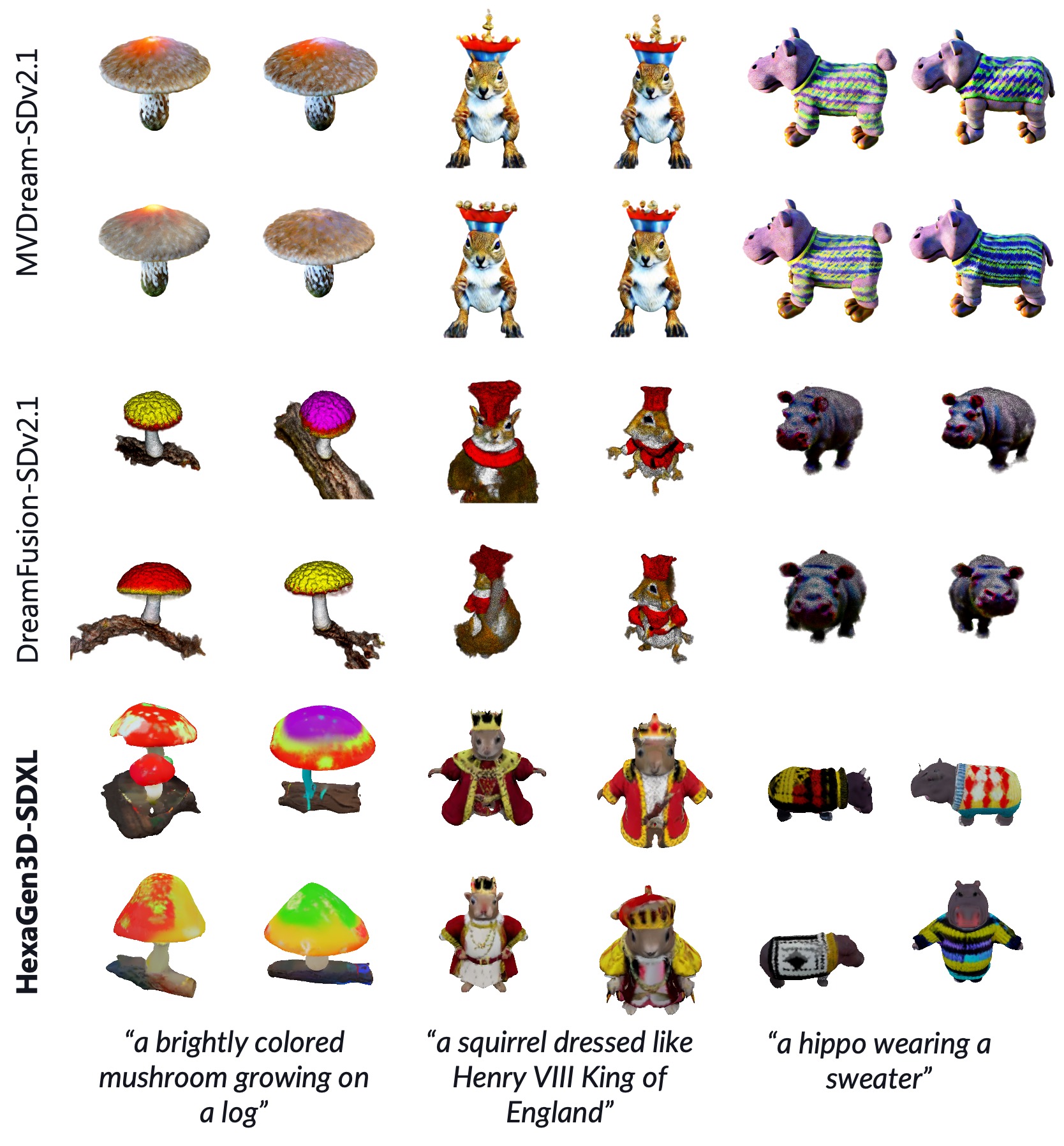}
        \caption{We show four (4) generations for each text prompt to demonstrate HexaGen3D generates more diverse 3D objects compared to the baseline MVDream-SDv2.1 and DreamFusion-SDv2.1 text-to-3D models.}
    \label{fig:qualt_comparison_diversity}
\end{figure}

\begin{table}[!t]
    \centering
    {\renewcommand{\arraystretch}{1.2}
        \small
        \begin{tabular}{l|c}
        \hline
        \textbf{Method} & \multicolumn{1}{c}{\textbf{CLIP score} $\uparrow$} \\ 
        \multicolumn{1}{c|}{} & CLIP L/14 
          
        \\
        \hline
        (A1) HexaGen3D-SDv1.5 
            & \cellcolor{tabthird}24.02 
            \\
        (A2) No Weight Sharing
            & 23.43 
            \\
        (A3) No Hexaview Baking 
            & 21.85 
            \\
        
        (A4) No Hexaview Prediction
            & 18.47 
            \\
        \hline
        (B1) HexaGen3D-SDXL
            &  \cellcolor{tabfirst}24.98 
            \\
        (B2) No Weight Sharing
            & \cellcolor{tabsecond}24.97 
            \\ 
        (B3) No Hexaview Baking
            & 23.59 
            \\
        \hline
        
        \end{tabular}
    }
    \caption{We quantitatively compare different design choices in our HexaGen3D model and provide more detailed descriptions in section \ref{sec:ablation_study}.} 
    \label{tab:quant_comparison_abl_stdy_1}
\end{table}

\subsection{Comparison against existing methods} 
\label{sec:comparison-against-existing-work}

\Cref{fig:qualt_comparison_text_to_3d} shows a comparison of HexaGen3D against Shap-E~\cite{jun2023shap},  DreamFusion~\cite{poole2022dreamfusion}, TextMesh~\cite{tsalicoglou2023textmesh}, and MVDream~\cite{shi2023mvdream}. Our approach demonstrates superior performance in most aspects, notably offering significantly faster run times. While MVDream excels in producing higher-quality meshes, the extensive generation time hinders its practicality, as this baseline takes up to 3 hours to create an object on an NVIDIA A100 GPU --- a process $1600\times$ slower than HexaGen3D, which accomplishes this in merely 7 seconds. Furthermore, MVDream exhibits limitations in terms of diversity, often leading to similar outputs across seeds. This point is illustrated in \cref{fig:qualt_comparison_diversity}, where we show 4 generations for the same prompt, varying the seed, emphasizing the advantage of HexaGen3D in generating more diverse meshes.

In our quantitative comparison (\cref{tab:quant_comparison_text_to_3d}), HexaGen3D, with variants based on SD v1.5 and SDXL, excels in mesh quality and prompt fidelity against these baselines, significantly outperforming Shape-E, the only other solution that can compete with HexaGen3D in terms of speed. Despite using an earlier version of StableDiffusion, our user studies and CLIP score evaluations indicate that HexaGen3D-SDv1.5 performs roughly on par with DreamFusion while being $180 \times$ faster. Switching to an SDXL backbone boosts our results significantly without impacting latency much. HexaGen3D-SDXL reaches a preference rate of $78 \%$ over DreamFusion (see \cref{fig:user_study}) and an average prompt fidelity preference score of $0.77$, narrowing the gap with MVDream's score of $0.88$ on this metric.

\begin{figure}[t!]
    \centering
        \includegraphics[width=0.9\linewidth]{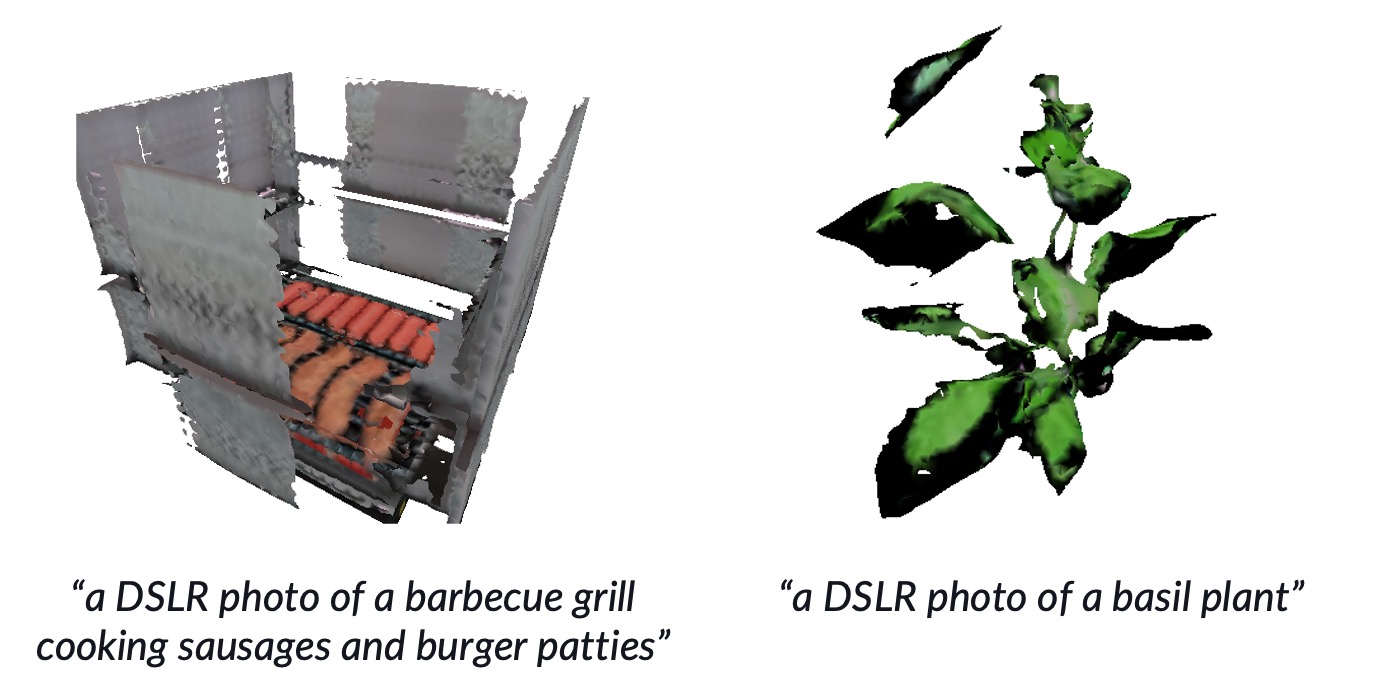}
        \caption{The generated 3D objects can contain minor limitations in terms of box artifacts, and intricate or thin mesh structures.}
    \label{fig:failure_cases}
\end{figure}

\subsection{Ablation studies}
\label{sec:ablation_study}

We comprehensively validate some of our design choices through ablation studies in Table~\ref{tab:quant_comparison_abl_stdy_1}. Key findings can be summarized as follows:

\begin{itemize}

\item \textbf{Multi-tasking multi-view and 3D generation is beneficial.} We find it interesting that the same U-Net can perform both hexaview diffusion and hexaview-to-triplanar mapping with reasonable accuracy. In rows A2 and B2, we study how utilizing two distinct U-Nets (same architecture, separate weights) for these two tasks affects performance. Interestingly, using two separate networks leads to a noticeable performance drop with SDv1.5 (A2). With SD-XL (B2), removing weight sharing does not make much of a difference --- but has the disadvantage of roughly doubling the model size and memory footprint.

\item \textbf{Hexaview baking provides a significant quality boost.} Removing the finer textural details produced by our baking procedure translates into significantly lower CLIP scores, both with SDv1.5 (A3) or SDXL (B3).

\item \textbf{Hexaview guidance is a necessary regularization.} Training the same model to directly denoise triplanar latents similar to 3DGen \cite{gupta20233dgen}, without hexaview guidance (A4), leads to significantly worsened performance, illustrating the importance of using six-sided multi-view prediction as a surrogate task to steer the model toward triplanar latent generation.

\end{itemize}

\section{Discussion}

Our results demonstrate that existing text-to-image models can be directly adapted to 3D asset generation with only minor modifications and minimal fine-tuning data. To facilitate this knowledge transfer, we introduce orthographic hexaview guidance, a surrogate task designed specifically to steer the 2D capabilities of existing text-to-image models toward 3D synthesis. This task, which amounts to predicting six-sided orthographic projections, can be thought of as a form of multi-task regularization. Additionally, we present simple yet effective architectural modifications that better fit existing 2D diffusion architectures for 3D generation. We show the effectiveness of our approach through extensive experimental results and believe these techniques are widely applicable to various text-to-image models. Notably, our approach scales well to larger pre-trained text-to-image models, with the SDXL variant significantly outperforming its SDv1.5 counterpart, both in terms of mesh quality and prompt fidelity. This scalability highlights HexaGen3D's potential in a context where 3D data is scarce.

 While relatively rare, 3D objects generated by HexaGen3D occasionally contain artifacts such as a box scaffolding enclosing the 3D object as shown in Fig.~\ref{fig:failure_cases} (left). This usually happens when the hexaview diffusion process generates inconsistent non-uniform backgrounds, leaving no choice but to add ``walls" around the main object during triplanar latent generation to resolve these inconsistencies. Additionally, as observed in many other 3D mesh reconstruction techniques, generating 3D objects with intricate or thin structures is notoriously challenging --- see  Fig.~\ref{fig:failure_cases} (right). Looking forward, exploring to what extent these issues could be resolved with more 3D data \cite{deitke2023objaverse-xl} would be interesting. Enhancing the VAE pipeline, which we mostly kept unchanged in this work compared to the original implementation proposed in 3DGen \cite{gupta20233dgen}, present another exciting avenue for future research.

In conclusion, we believe HexaGen3D stands out as an innovative and practical solution in the field of 3D asset generation, offering a fast and efficient alternative to the current generation methods. HexaGen3D generates high-quality and diverse textured meshes in 7 seconds on an NVIDIA A100 GPU, making it orders of magnitude faster than existing approaches based on per-sample optimization. Leveraging models pre-trained on large-scale image datasets, HexaGen3D can handle a broad range of textual prompts, including objects or object compositions unseen during finetuning (see supplementary materials for more results, including generations from MS-COCO captions). Furthermore, HexaGen3D generates diverse meshes across different seeds, a significant advantage over SDS-based approaches like DreamFusion or MVDream. We believe these attributes make HexaGen3D a pioneering tool in the rapidly evolving domain of 3D content creation.

{\small
\bibliographystyle{ieee_fullname}
\bibliography{PaperForReview}
}

\end{document}